\documentclass[letterpaper, 10 pt, conference]{ieeeconf}  

\IEEEoverridecommandlockouts                              

\usepackage{graphicx}
\usepackage{xcolor}
\usepackage{amsmath} 
\usepackage{placeins}  
\usepackage{calc}   
\usepackage{caption}
\usepackage{subcaption} 
\usepackage{lastpage}
\usepackage{multirow}
\usepackage{hyperref}
\usepackage[normalem]{ulem}

\renewcommand{\arraystretch}{1.3}

\newcommand{\brox}[1]{\textcolor{purple}{#1}}

\usepackage{rotating}
\usepackage{accents}
\usepackage{amsfonts}
\usepackage{pifont}
\newcommand{\xmark}{\ding{55}}%
\newcommand{\tmark}{$\sim$}
\newcommand{\cmark}{\checkmark}
\newcommand{\myvec}[1]{\accentset{\rightharpoonup}{{#1}}}
\newcommand{\Max}[1]{\textcolor{red}{#1}}
\newcommand{\Evan}[1]{\textcolor{blue}{#1}}

\title{\LARGE \bf
Conditional Visual Servoing for Multi-Step Tasks
}

\author{Sergio Izquierdo$^*$, Max Argus$^*$, Thomas Brox
\thanks{* equal contribution authors}
\thanks{All authors are at the University of Freiburg and are members of Brain-Links-Brain-Tools. The corresponding author can be contacted via: {\tt\small argusm@cs.uni-freiburg.de}}%
}

\begin{document}

\maketitle
\thispagestyle{empty}
\pagestyle{empty}

\begin{abstract}


Visual Servoing has been effectively used to move a robot into specific target locations or to track a recorded demonstration. It does not require manual programming, but it is typically limited to settings where one demonstration maps to one environment state. We propose a modular approach to extend visual servoing to scenarios with multiple demonstration sequences. We call this conditional servoing, as we choose the next demonstration conditioned on the observation of the robot. This method presents an appealing strategy to tackle multi-step problems, as individual demonstrations can be combined flexibly into a control policy.
We propose different selection functions and compare them on a shape-sorting task in simulation. With the reprojection error yielding the best overall results, we implement this selection function on a real robot and show the efficacy of the proposed conditional servoing. For videos of our experiments, please check out our project page: \href{https://lmb.informatik.uni-freiburg.de/projects/conditional_servoing/}{\color{blue}https://lmb.informatik.uni-freiburg.de/projects/conditional\_servoing/}
\end{abstract}

\vspace{1em}
\section{INTRODUCTION}

Programming robots to complete tasks remains a challenging and time-consuming problem, which prevents robots from being used more widely. 
Servoing-from-Demonstration (SfD) approaches enable few-shot imitation from demonstration videos and, thus, are an attractive direction for robot programming. The key challenge is how to make SfD approaches flexible enough to automatically adapt to many variations of a task without the need to record a demonstration for each particular case. 

Our work builds on FlowControl~\cite{Argus2020}, which servos to the configuration specified in a demonstration frame, then iterating to the next demonstration frame once one is close enough. This process is repeated for all demonstration frames, enabling the replication of demonstration tasks. The algorithm requires RGB-D observations as well as foreground object segmentations for the demonstrations.
Robustness to variations in appearance is achieved by establishing correspondence between observation and demonstration via learned optical flow.  

FlowControl is bound to a single demonstration that must match the control problem. In this paper, we allow for a set of demonstrations to choose from. This enables the option to select the most suitable demonstration, conditioned on the initial demonstration frame, which is why we call the approach conditional servoing. Moreover, it enables the recombination of subtask demonstrations, similar to skills, to more complex policies and avoids the necessity of a complete demonstration for each possible permutation of the scene.

\begin{figure}[!t]
\centering
 \centering
 \includegraphics[width=\linewidth]{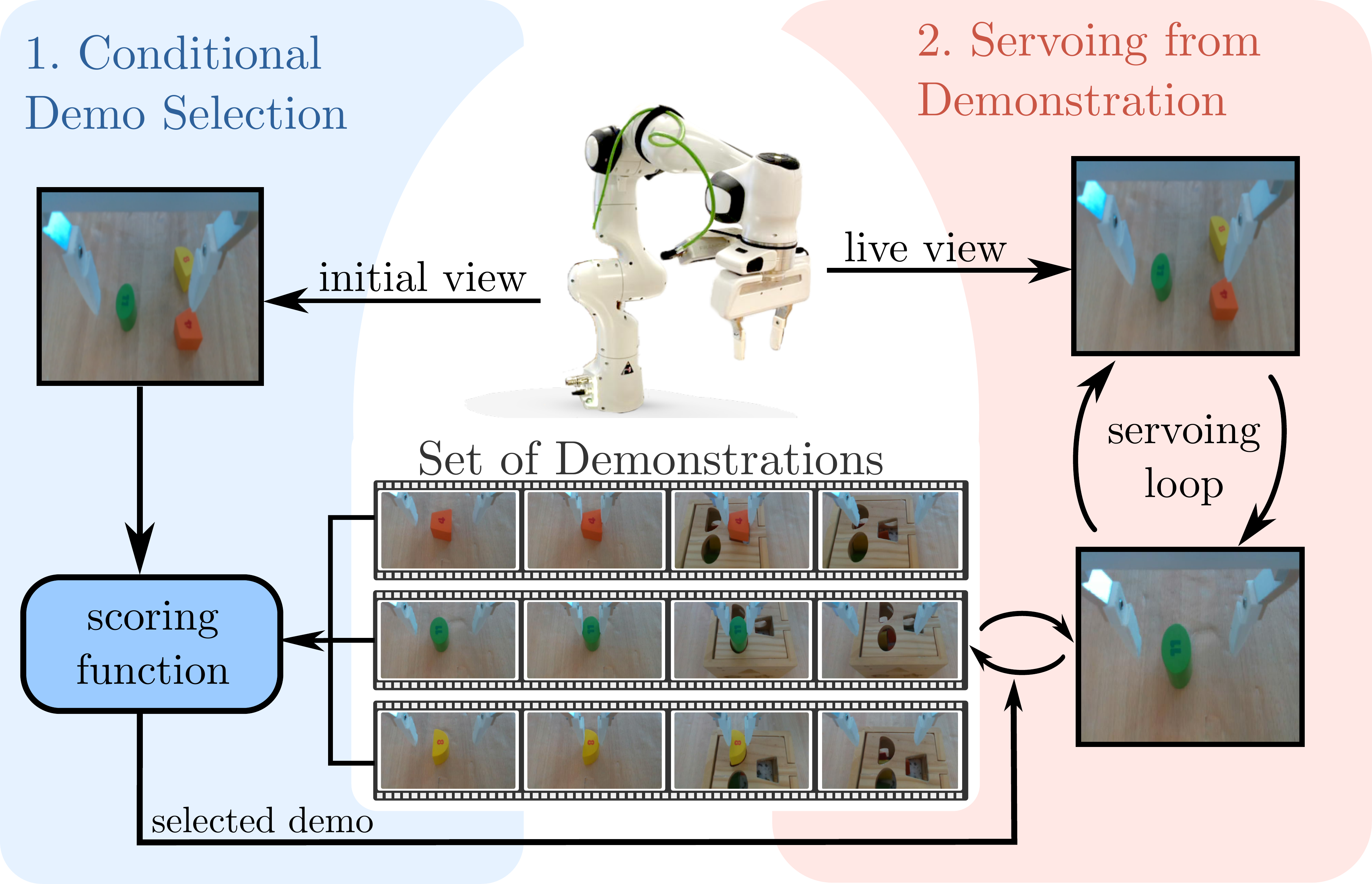}
 \vspace{.3em}
 \caption{In conditional servoing, the most suitable demonstration is selected. In our approach, this is done by evaluating a distance score between initial live views and initial demonstrations views, shown in the blue half. This demonstration is then used with a servoing algorithm designed to replicate demonstration sequences to complete tasks, shown in the red half. A more detailed description of the method is shown in Figure \ref{fig:method}. The images are from an end-of-arm camera, which we use for robot control.}
 \label{fig:teaser}
\end{figure}

More precisely, the system is given a set of demonstrations of different subtasks, some of which may be unsuitable to follow, for instance, the object is not currently visible. In conditional servoing, the system must select the best demonstration to follow based on its current observation. To do so, we propose to perform the selection based on a matching score that assesses the similarity between the demonstration and the current observation

We evaluated a number of different matching scores in simulation and verified that the most promising scoring metric works well also on a real robot. Our experiments in simulation and on the real robot deal with a shape sorting task in which multiple blocks are present in the scene and must be inserted in the right opening of a sorter box. This task is challenging from the control point of view, due to the tight holes of the sorter box, and requires the selection of the right entry point. Conditional servoing increases the success rate of FlowControl on the real robot from 40\% to 78\%. Furthermore, conditional servoing significantly increases the chances of correctly recovering by re-grasping if an object is dropped by mistake. In addition, we perform a qualitative evaluation on a number of related configurations.

\section{RELATED WORK}
\label{sec:related_work}


\textbf{Servoing:}
Visual Servoing \cite{Kragic02, Chaumette06} is a well-established approach to vision-based robotics problems, in which a comparison between sensor observations and a target view, or demonstration, is used to generate control commands in a closed feedback loop.
This allows the control to adapt to the surrounding environment as observed by the system. A large number of robotics problems can be approached with visual servoing techniques \cite{Hutchinson96}, examples of this include manufacturing \cite{Kheddar19} and surgery \cite{Huang17}, among many others.

While we did not find any prior works featuring demonstration selection in visual servoing, there are some related concepts. In hybrid visual servoing, control laws are extended with additional constraints \cite{Chaumette16}.
This allows the definition of different subtasks, e.g., obstacle avoidance, manipulation tasks, and joint-limits avoidance \cite{Lippiello16}. Subtasks are usually defined simultaneously and often pertain to some constraining of the robot state.
Task sequencing in visual servoing describes a situation, where the control laws are changed over the course of iteration \cite{Mansard2004}.
In contrast to these works, we use the term \emph{tasks} to describe the problems in the environment that the robot should complete. 

Our work extends FlowControl~\cite{Argus2020}, which combines the use of learned optical flow with demonstration sequence tracking to replicate tasks.
The use of learned optical flow for servoing is similar to \cite{Nelson93, Mitsuda99, Sepp2006}, and the use of demonstration sequence tracking is similar to \cite{ Sermanet17, Huang17}.
For the sake of completeness, we briefly recapitulate how FlowControl works. It has two main parts, a frame alignment part and a sequence tracking part.
The frame alignment part computes the transformation between the live state, as seen by the robot's camera, and the state shown in the selected demonstration frame. This transformation is used to move the robot accordingly. The method relies on optical flow estimates to obtain correspondences between both states.
The correspondences are further filtered using a foreground segmentation (attention mask) of the demonstration before being converted into 3D points using the camera's depth measurements. Finally, a rigid 3D transformation is computed between the point clouds.
After achieving a good alignment with the current frame, the method performs sequence tracking by iterating to the next demonstration frame, allowing for the replication of tasks. A visual explanation of the method is available on Youtube: \url{https://youtu.be/SSC8RyBGUug}.

\vspace{.5em}
\textbf{Compositionality:}
Individual demonstrations can be seen as encoding different skills.
A large number of robotics works investigate compositional skills. These include \cite{Kroemer2021}, associative skill memories \cite{Pastor2012}, as well as the closely related hierarchical approaches \cite{Kroemer2015}.
More recent works have extended this line of work to the deep-learning domain \cite{Shu2018, Mees2020}.

\vspace{.5em}
\textbf{Frame Selection}:
Selecting a suitable demonstration sequence can be addressed as a registration problem, in which frames from demonstrations are registered to a live view. A large number of similar problems exist in computer vision.
Image retrieval tries to find the most similar image from a database of candidates \cite{Chum09, Jgou10, Jun19}.
It is the basis of visual place recognition, where one tries to localize based on a query photograph \cite{Arandjelovi18, Weyand20}.
Some early works using visual servoing can be regarded as predecessors of SLAM \cite{Rasmussen1996}. More recent incarnations of similar ideas can be found, for example, in \cite{Delfin2014}.
In the field of SLAM, similar problems are encountered in global localization \cite{Lim12} or re-localization \cite{Williams11}, which is closely related to the problem of loop closure.

Registering images to a map is also encountered in Structure-from-Motion (SfM), in which large numbers of images are registered to a scene \cite{Schoenberger16}.

The registration methods described here differ substantially in the representations they compare. SLAM methods often compare sequences of observations to maps. We are interested in comparing single images to single images. Furthermore, we are interested in just the registration of the relevant foreground objects. Thus, these methods cannot be used in our approach out-of-the-box. A similar setting is encountered in the context of object pose estimation. There is also the problem of template-based object instance detection \cite{Mercier21, Rad17}, in which objects need to be localized within a scene.


\begin{figure*}[!t]
\centering
 \centering
 \includegraphics[width=.98\linewidth]{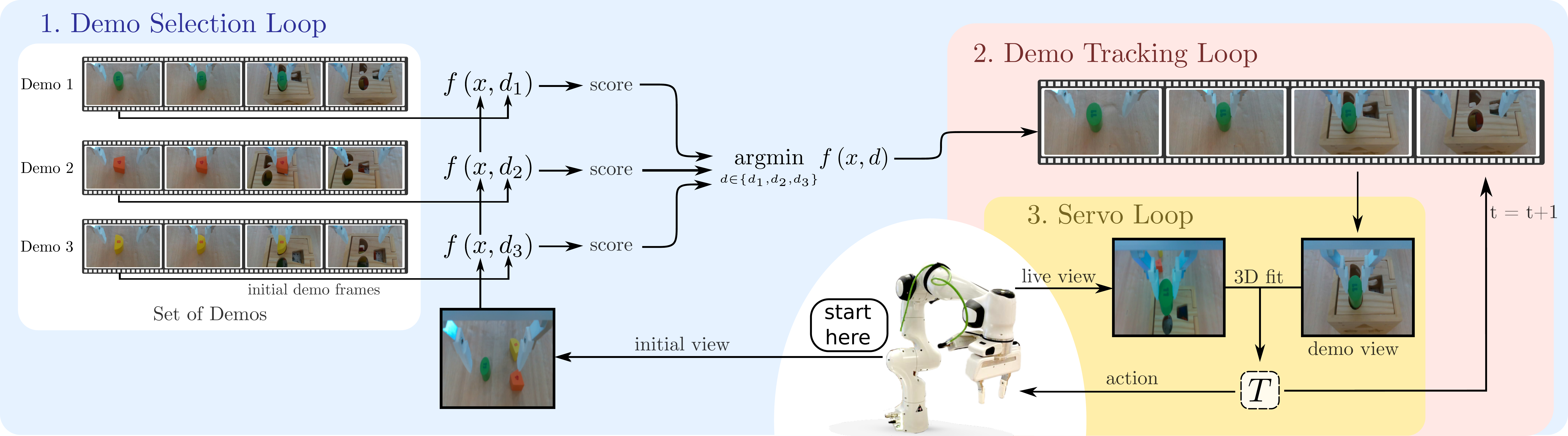}
 \caption{The demonstration selection, shown in blue, is performed by taking the first frame from a  demonstration sequence and comparing it to the initial live view. We then select the closest demonstration and continue with servoing from a demonstration sequence using FlowControl, shown in red and yellow. Once a demonstration sequence is done, demonstration selection can be repeated until the overall task is completed.}
 \label{fig:method}
\end{figure*}

\vspace{1em}
\section{METHOD}
\label{sec:method}

This work considers a setting where multiple demonstration sequences are available and a suitable demonstration must be chosen. We call this conditional servoing because our choice of sequence depends on observations of the current environment.

Our setting implies environments in which correspondences between the current state and some of the demonstrations may be poor or even nonexistent.
Hence, we investigate selection mechanisms to retrieve the most suitable demonstration from a set of multiple demonstration sequence options. This allows following the selected demonstration and discarding unsuitable ones.

This selection is implemented as an outer loop around a servoing algorithm.
The loop starts by selecting the best demonstration and replicating it. Once it is completed, the next demonstration can be selected and replicated. This allows addressing multi-step tasks. A visual depiction of our method can be seen in Figure \ref{fig:method}.

In the following, we describe the setting formally.
We want to select the best demonstration, $D_{best}$, from a set of demonstrations, $\mathcal{D}=\{D_1, D_2, ..., D_n\}$. Given a demonstration $D_i$, our selection is based on only $d_i$, its first frame, as well as the current observation of the robot, $x$. To do so, we employ a distance function to assign a score to each demonstration, based on the similarity to the current observation, $f\left(x, d\right)$ 

As we build on FlowControl as the Servoing-from-Demonstration method in the inner loop, we can start with evaluating the residual of the rigid-transformation estimation to assess the quality and reliability of the matching. For a more detailed description of FlowControl, please see Section \ref{sec:related_work}.



We propose four different distances to assess the quality of a given demonstration:

\vspace{.8em}
\textbf{Point-Quality Distance:}
We consider using the estimated 3D transformation between the live and the demo's initial point clouds, $p^L, p^D$. We transform the live point cloud into the demo's one according to the transformation retrieved by FlowControl, $(R, t)$. After this, outliers that are more than 5mm apart are removed. Finally, the L2 norm between the two point clouds is computed to obtain the distance.

\begin{equation}
   D_\text{PQ} = \frac{\sum_{i \in \Omega_p} \parallel(R \cdot p_i^D + t) - p_i^L\parallel}{|~\Omega_p|}
\end{equation}

\vspace{.8em}
\textbf{Color-Quality Distance:}
Instead of comparing the 3D positions of the points, in this distance, we focus on the points' colors, $C(p)$. The L2 distance between the RGB values of the point cloud correspondences is calculated. This distance aims to detect situations where the point clouds obtain a good registration, but the colors of the object in the live state and in the considered demonstration do not match.
As in the previous distance, possible outliers have been discarded, and are not considered in the distance.


\begin{equation}
  D_\text{CQ} = \frac{\sum_{i \in \Omega_p} \parallel C^{D}(p_i^D) - C^{L}(p_i^L)\parallel}{|~\Omega_p|}
\end{equation}

\vspace{.8em}
\textbf{Reprojection Distance:}
Similar to the color-quality distance, the reprojection distance evaluates the accuracy of the RGB correspondences on the basis of images. As point correspondences between the current observation and the demonstration frames are obtained with an optical flow estimation, we propose to evaluate the suitability of the demonstration by evaluating the quality of the flow. As we, naturally, lack the optical flow ground truth, we follow the approach of some self-supervised flow estimation methods: the use of photometric reconstruction as the evaluation function~\cite{jason2016back, ren2017unsupervised}. 

\begin{figure}[!htb]
\centering
\begin{tabular}{@{}c@{}c@{}c@{}}
 \includegraphics[width=.33\linewidth]{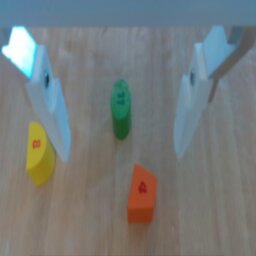} &
 \includegraphics[width=.33\linewidth]{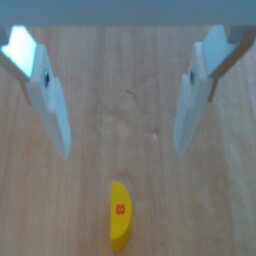} &
 \includegraphics[width=.33\linewidth]{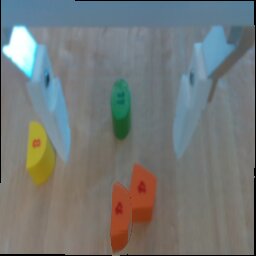} \\
 \includegraphics[width=.33\linewidth]{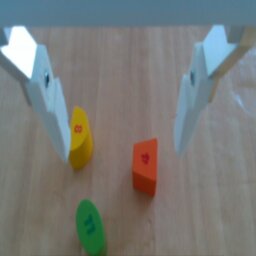} &
 \includegraphics[width=.33\linewidth]{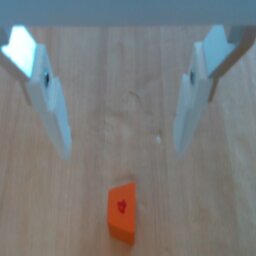} &
 \includegraphics[width=.33\linewidth]{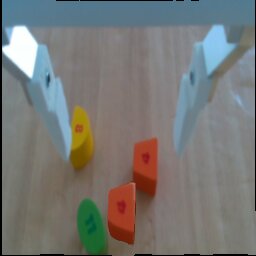} \\
live view & demo view     & re-projection \\ 
$I^L$ & $I^D$     & $I^L(x+u,\, y+v)$ \\ 

\end{tabular}
 \caption{Reprojection examples. Optical flow is computed from the demo view (center) to the live view (left). Using this flow, we warp the live view to be similar to the demonstration view. The figure shows a bad warping in the top row, where the flow matched the yellow object with the orange one, as well as a good warping in the bottom row.}
 \label{fig:reprojection} 
\end{figure}

The live view, $I^L$, is warped into the demo view, $I^D$, by using the optical flow, $u, v$, between the demo view and the live view. By computing the dissimilarity between the warped (reconstructed view) and the demo, we obtain a measurement of the fit of the flow. We use the L2 distance between the two, and only calculate the dissimilarity in the pixels selected by the demonstration's foreground mask, $\Omega_M$, as other pixels do not need to be estimated correctly. An example of a reprojection image is shown in Figure \ref{fig:reprojection}.

In contrast to the previous functions, this one relies only on two-dimensional images. Moreover, it considers all pixels in the segmentation mask, i.e., it does not discard possible outliers. 
Outlier removal can have both benefits and drawbacks, while it can make the distances more robust it can also be problematic to apply in situations where the objects are not visible in the current state, which can result in incorrect transformation estimations. 

\begin{equation}
    D_\text{RP} = \frac{\sum_{x,\, y\,\in\,\Omega_M}{\parallel I^L(x+u,\, y+v) - I ^D(x,\, y)\parallel }}{|~\Omega_M|}
\end{equation}

\vspace{.5em}
\textbf{Learned distance:}
Trying to find interactions between the previously presented distances, we propose to train a classifier to predict the probability of success of a given pair $(x, d)$. We train a small multilayer perceptron that takes the other distances as inputs and predicts the probability of success. The best demonstration is selected by taking the one with the highest estimated probability of success.

We train a small MLP with three hidden layers of $\theta=64$, $64$, and $32$ neurons. The last layer has one neuron to predict the probability of success. We use ReLU activation in all intermediate layers and logistic in the last one. For the loss the binary cross-entropy is used and optimized with \textit{Adam} \cite{kingma2014adam} using its default recommended parameters of $\beta_1=0.9$ and $\beta_2=0.999$ as well as an initial learning rate of $10^{-5}$. We used a batch size of 128 and trained for $2500$ iterations or until the learning has converged.

\begin{equation}
	D_\text{MLP} = 1 - P_\text{MLP}(y\mid\,x, \,d, \,\theta)
\end{equation}


\vspace{.5em}
\section{EXPERIMENTS}


We use a shape sorting task to evaluate the conditional servoing. It is a popular multi-step task also used in other works \cite{Levine15, Ibarz2021}.
The robot has to grasp three pieces and introduce them precisely on a shape sorting cube, one after the other. To do so, the robot has to select the best demonstration to follow before grasping the next piece. Given the small tolerances, the task requires very precise grasping, positioning, and insertion of the objects. This is much easier if the optimal demonstration is selected. 

\subsection{Experimental Setup}

\begin{figure}[!tb]
\centering
 \includegraphics[width=.49\linewidth]{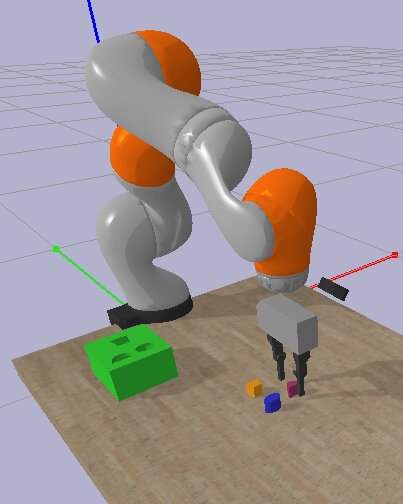}
 \includegraphics[width=.49\linewidth]{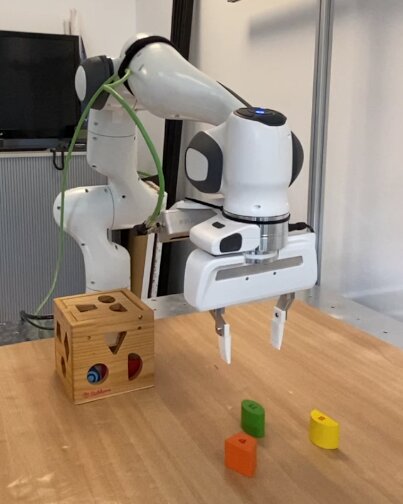}
 \caption{Robot setup with end-of-arm camera in simulation (left) and in the real environment (right).}
 \label{fig:setup}
\end{figure}

\begin{figure}[!htb]
\centering
 \includegraphics[width=.45\linewidth]{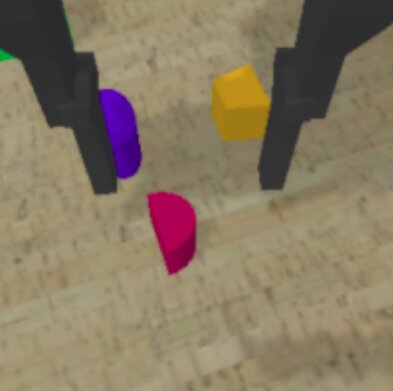}
 \includegraphics[width=.45\linewidth]{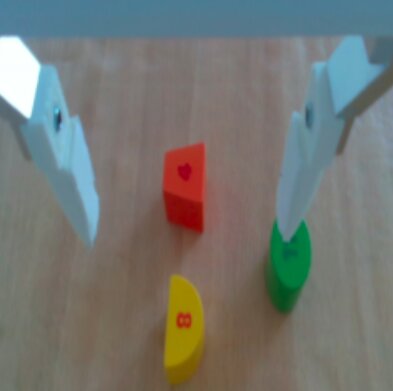}
 \caption{Simulation environment (left) used to better perform large-scale quantitative evaluation of selection scoring functions compared to the real environment (right).}
 \label{fig:simandreal}
\end{figure}

We re-created the physical shape sorting task in the PyBullet physics simulator~\cite{Coumans2021} to afford to collect a large number of executions, something prohibitively expensive in a real robot\footnote{Note that the simulation was only used for additional evaluation. In contrast to many contemporary robot learning methods, the simulation is not needed to learn the task. The real robot can be instructed by demonstration of the subtasks in the physical world.}. As shown in Figure~\ref{fig:simandreal}, we aligned the simulation and the real robot setup, but the simulation uses the Kuka iiwa robot, whereas the physical robot was a Franka Emika Panda robot. It was equipped with an Intel RealSense D435 RGB-D camera mounted on the robot's end-effector, obtaining an eye-in-hand setup. For the gripper, two 3D printed fingers are attached to the robot's fingers. This setup is shown in Figure~\ref{fig:setup}.

\vspace{1em}

For the control, we restricted the end-effector's freedom such that it always faces downwards, limiting its orientation to the yaw rotation. We parametrize the control as in~\cite{Argus2020}, with a 5DoF continuous action $a=[\Delta x, \Delta y, \Delta z, \Delta \theta, a_{\text{gripper}}]$ in the end-effector frame. $ \Delta x, \Delta y, \Delta z$ specify a Cartesian offset for the desired end-effector position, $\Delta \theta$ defines the yaw rotation of the end effector, and $a_{\text{gripper}}$ is the gripper action that is mapped to the binary command to open or close the fingers. 


\vspace{.5em}
\subsection{Simulation Experiments}

First, we evaluated the scoring functions defined above. Because the simulation allows for effortless larger-scale experiments, we ran this evaluation in simulation. 
As the MLP distance function requires training, we recorded $500$ executions with three objects each, giving a total of $1500$ runs. Of these, we use $1000$ samples to learn the distance and evaluate on the remaining samples as a test set. Details of the network's architecture and the training process are described in Section \ref{sec:method}.

The suitability of a demonstration can only be evaluated by tracking it with visual servoing to completion and seeing if the task is successfully completed. Thus, to evaluate a proposed scoring function, we run $100$ simulations with each of the three objects, resulting in $300$ samples, selecting the optimal demonstration according to the given scoring function. The starting arrangement of the objects in each of the $100$ simulations is randomized, but the scoring functions were evaluated with the same set of initial configurations to enable a fair evaluation of their effectiveness.

Table~\ref{tab:success_probs} shows the success rate obtained in simulation when following the proposed distances. We separate the results considering \textit{all objects}, counting all recorded samples, and \textit{first object}, only considering runs inserting the first of the three objects. This distinction shows the benefits of our conditional servoing in multimodal situations, where the flow estimation can match incorrect objects with the demonstrations. The point cloud based methods were also tested without outlier removal, which did not improve results.

\begin{table}
\begin{center}
\caption{Simulation success rates for different selection scores evaluated on the simulated shape sorting task.}
\begin{tabular}{l|r|r}
Score for argmin sampling& \multicolumn{2}{c}{Success Rates [\%]}\\
of demonstration sequence& First Object & All Objects \\ \hline \hline
Uniform Random  &   55     &  47  \\ \hline
Point-Quality  &   69     &  60  \\ \hline
Color-Quality  &   76     &  58  \\ \hline
Reprojection  &   81     &  63  \\ \hline
MLP  &   81     &  65     \\ \hline
\end{tabular}
\label{tab:success_probs}
\end{center}
\end{table}

All our proposed distances obtain a higher success rate than the uniform random baseline, where demonstrations are picked according to a uniform random distribution. Of these, the reprojection and MLP distance yield similar success rates for the \textit{first object}. 
The MLP distance achieves slightly higher performance on the \textit{all objects} set, as it was trained on this quantity. However, we still chose the reprojection distance as our preferred method, as it does not require collecting a dataset and training.


Although the color-quality and the point-quality distance may seem more robust, as they discard possible outliers, they achieve a lower success rate than the reprojection distance. We argue that incorrectly computed transformations in the servoing may lead these functions into inconsistent results when the fitting is based on incorrect flow estimates. On the other hand, the reprojection distance directly evaluates the quality of the flow prediction.

\begin{figure}
\centering
 \includegraphics[width=.98\linewidth]{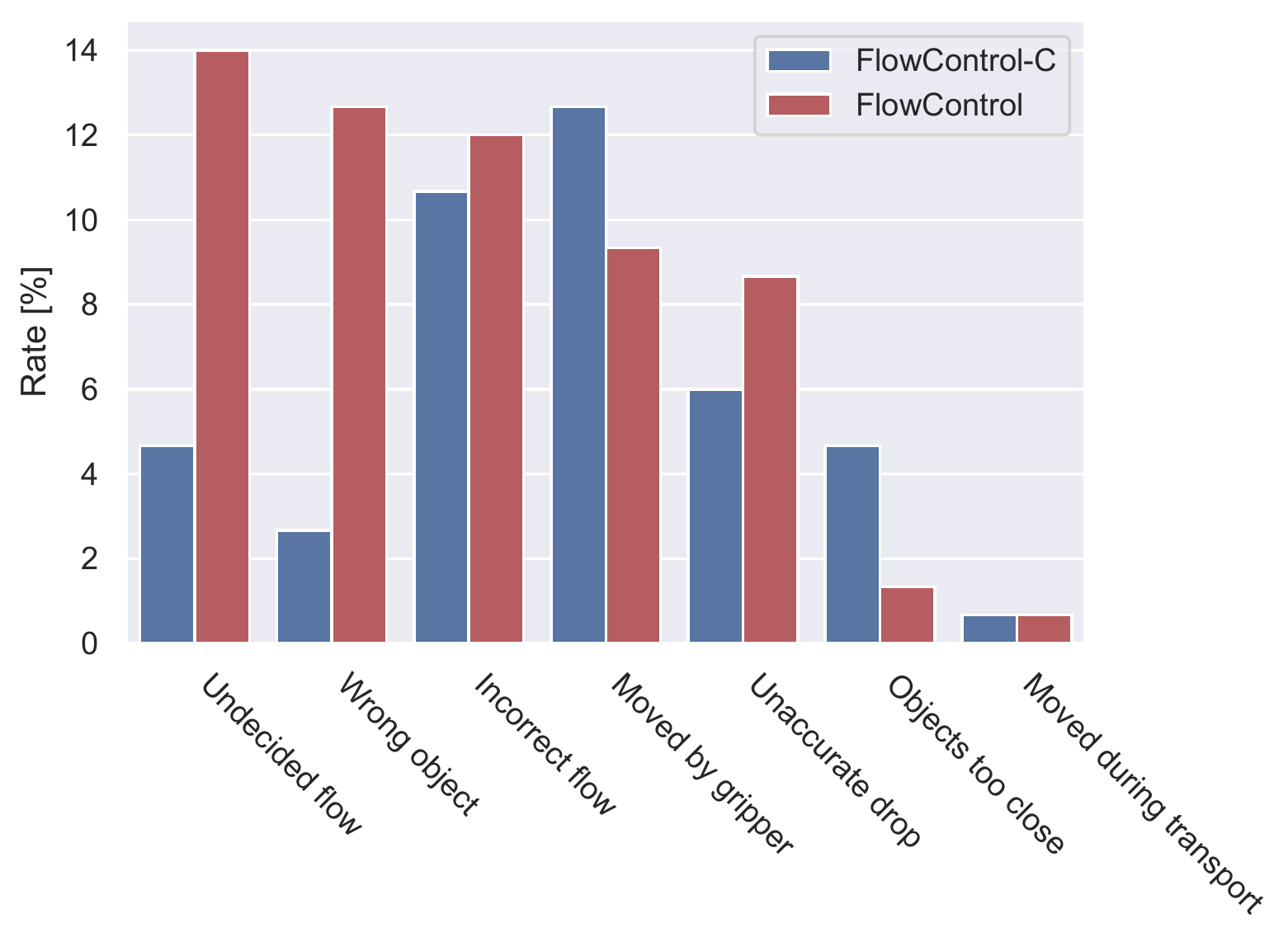}
 \caption{Error analysis for simulation experiments. Error types were assessed manually by observing a set of 60 and 88 failure runs in FlowControl-C and FlowControl respectively. Our method effectively reduces the top types of error of the baseline method.}
 \label{fig:barplot}
\end{figure}

Figure~\ref{fig:barplot} shows a detailed manual analysis of a set of 150 simulation runs. For these, we categorize the main cause of failures for the baseline FlowControl method, as well as our conditional servoing method, which uses reprojection distance as selection criterion. This is named FlowControl-C.
The three main causes of error in FlowControl stem from wrong optical flow estimates. In 
\textit{undecided flow} the flow estimation is unstable and jumps between different objects in each step. In \textit{wrong object} cases, the robot grasps an incorrect shape due to a wrong projection, and finally, in \textit{incorrect flow}, estimates are not accurate and, thus, the servoing does not converge or converges to an incorrect robot position. Including our conditional servoing significantly decreases these errors, showing one of the main benefits of our approach: an improvement in correspondence computation.

In an additional experiment, we show that the conditional servoing approach can also help when recovering from failure cases. To evaluate this in simulation we generate failure cases by opening the gripper according to a Bernoulli distribution with a probability of $p = 0.25$. To evaluate recovery we compare to the baseline of tracking the previously followed trajectory, which results in a success rate of 27\%. This is clearly outperformed by the conditional selection of the trajectory which has a success rate of 59\%.



\vspace{.5em}
\subsection{Real Experiments}

To test the effectiveness of our proposed conditional servoing we use a shape sorting task on the real robot. Based on the simulation experiments we selected the reprojection distance as the best overall selection score and subsequently carried out all our real robot experiments with this choice. These experiments are named FlowControl-C. An example of reprojected images is given in Figure \ref{fig:reprojection}.
We then performed quantitative and qualitative experiments with our new method.


\vspace{.5em}
\textbf{Success Rate:} 
Similar to experiments in simulation we start by evaluating the success rate achieved on the shape sorting task when using the baseline FlowControl method compared to our FlowControl-C method with the reprojection distance. The shape sorting task is evaluated using the \textit{first object}, as we wanted to test the method in multi-modal scenarios. With this setup, the flow estimation module is more likely to produce erroneous results due to multiple objects laying on the scene.
An example of this possible failure case is shown in Figure~\ref{fig:reprojection} top row.

The results of this experiment can be seen in Table \ref{tab:sr_real}. We obtain a similar improvement rate as in the simulation, confirming that a guided selection of the demonstration significantly increases the probability of success.

\begin{table}[tb]
\caption{Success rates for the shape insertion task evaluated in simulation and on the real robot. Compared to the baseline FlowControl method, success rates on both tasks increase when using our improved conditional servoing method called FlowControl-C.
}
\label{tab:sr_real}
\begin{center}
\begin{tabular}{c||c|c|}
&Simulation & Real Robot \\
\hline
FlowControl   & 55\% & 40\% \\
\hline
FlowControl-C & 81\% & 78\% \\
\hline
\end{tabular}
\end{center}
\end{table}

\begin{figure}[!htb]
\centering
 \includegraphics[width=.24\linewidth]{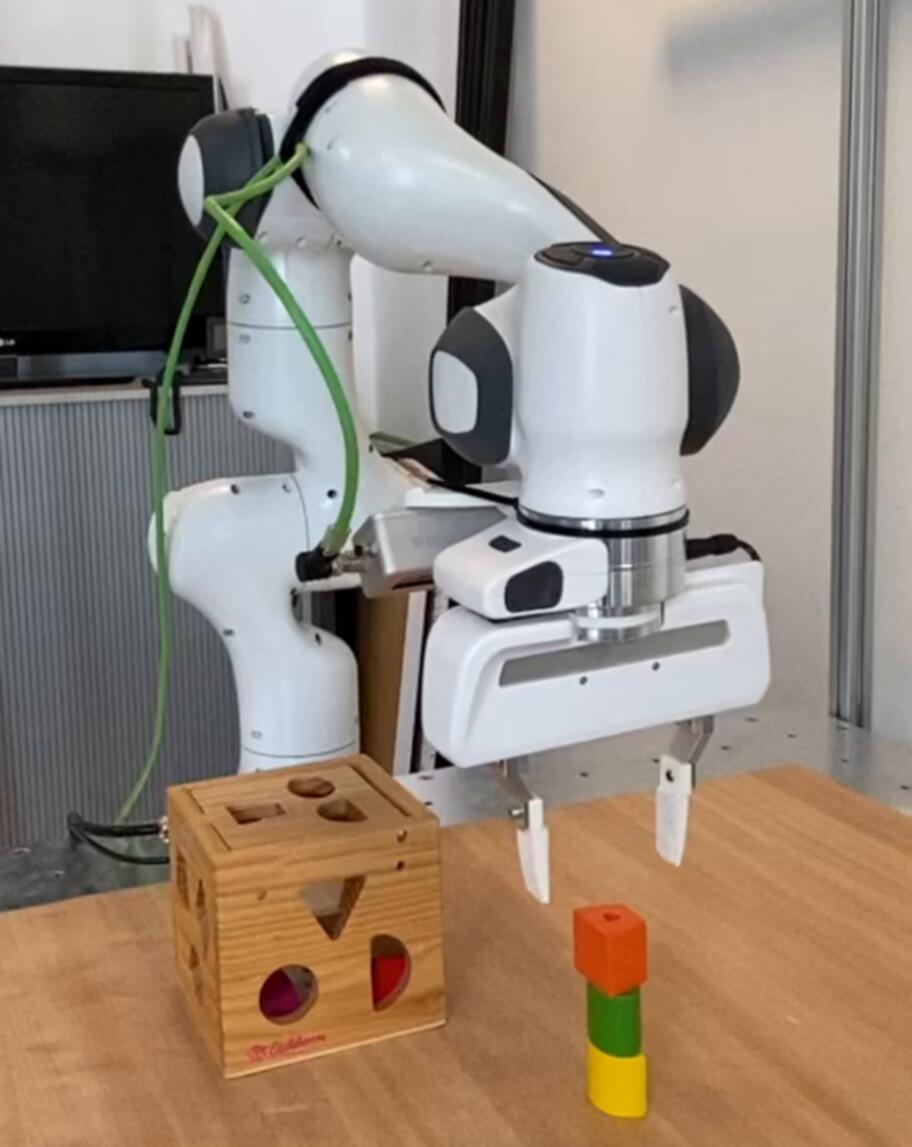}
 \includegraphics[width=.24\linewidth]{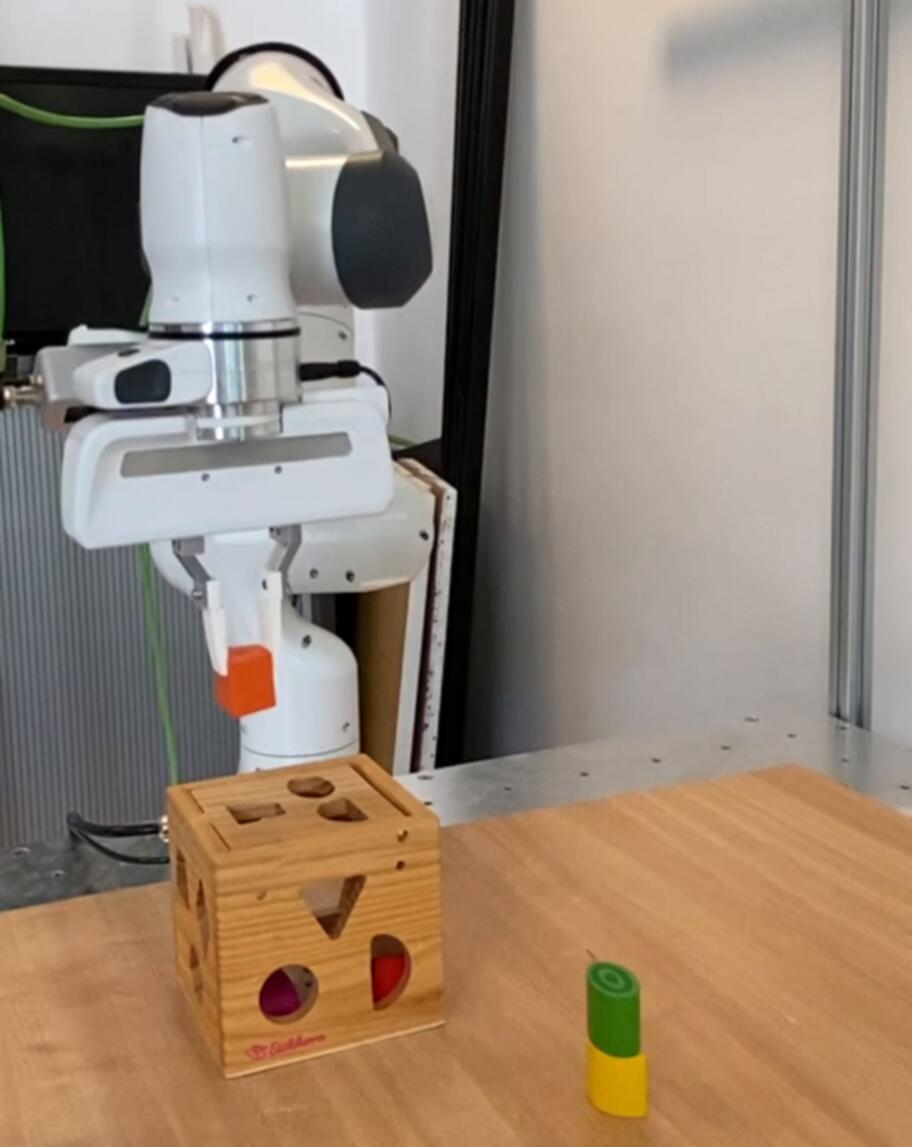}
 \includegraphics[width=.24\linewidth]{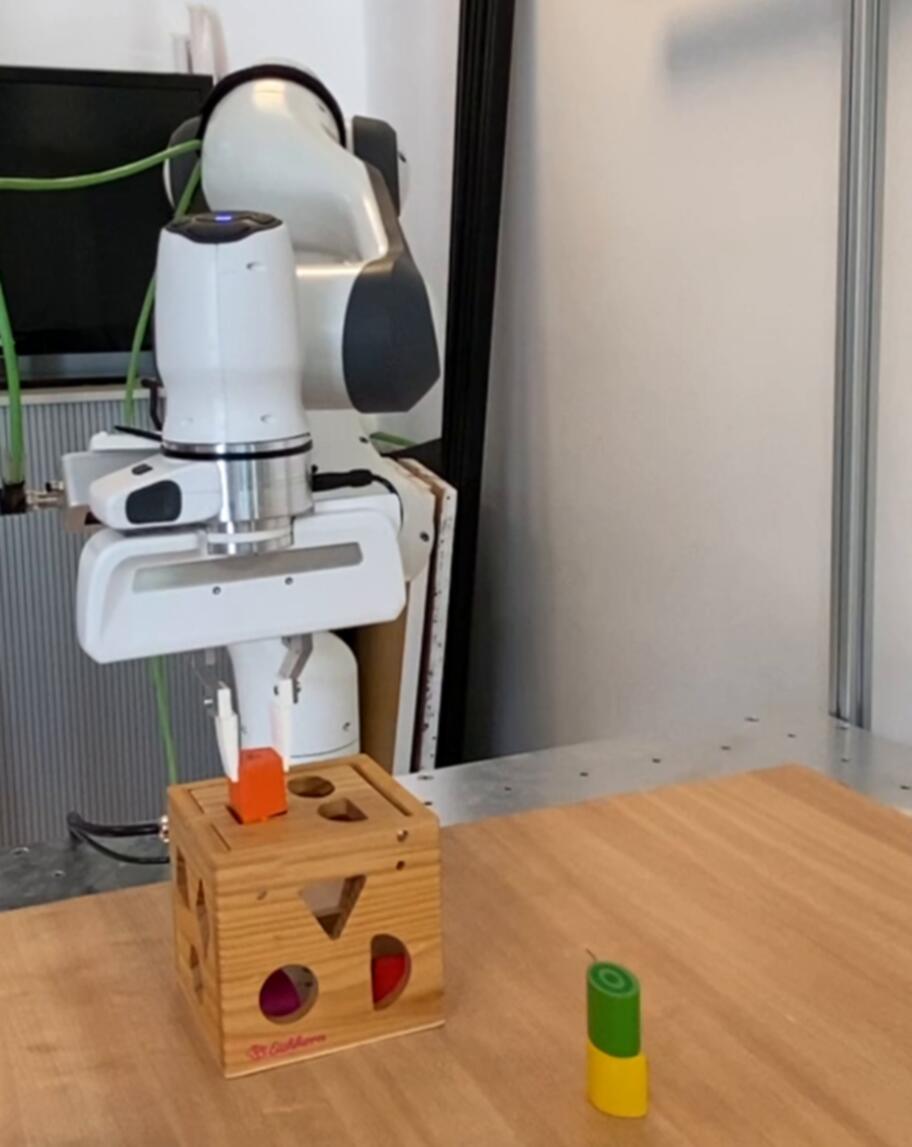}
 \includegraphics[width=.24\linewidth]{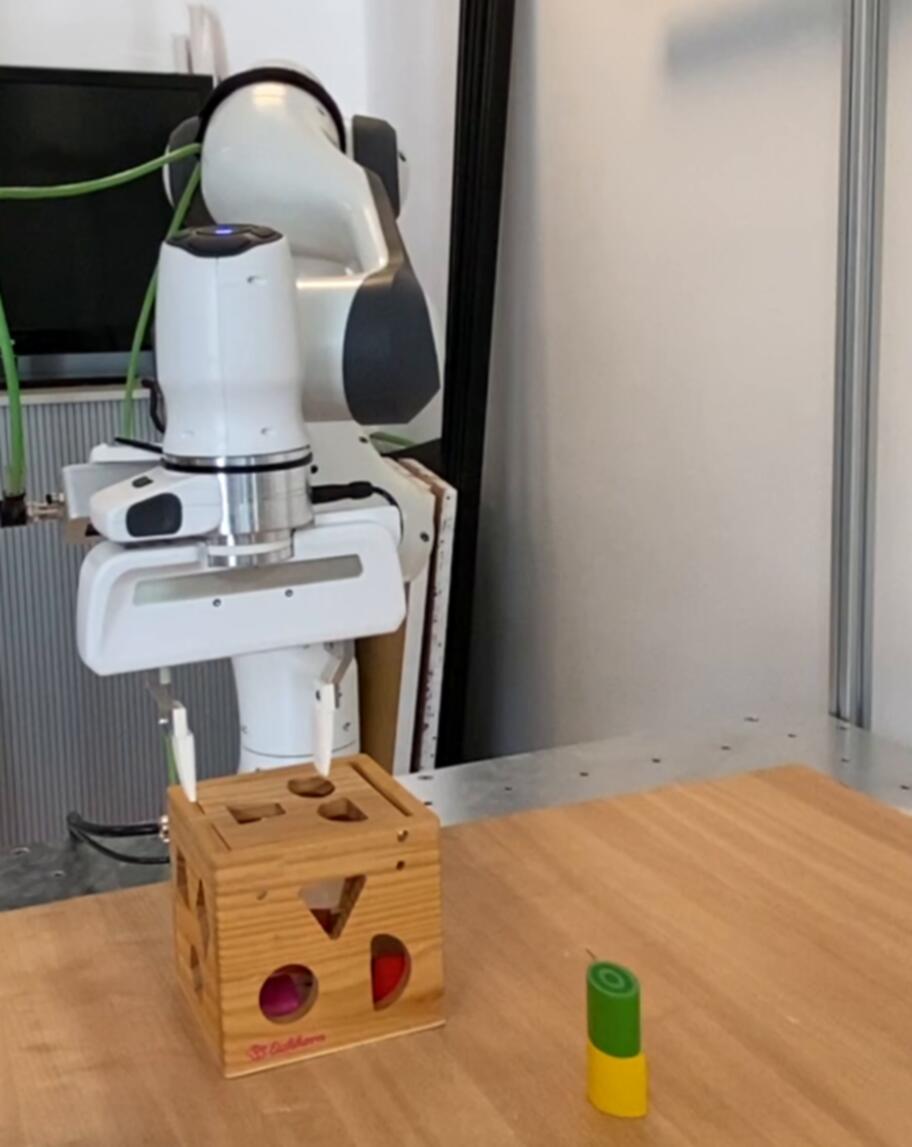}\vspace{0.5em}
 \includegraphics[width=.24\linewidth]{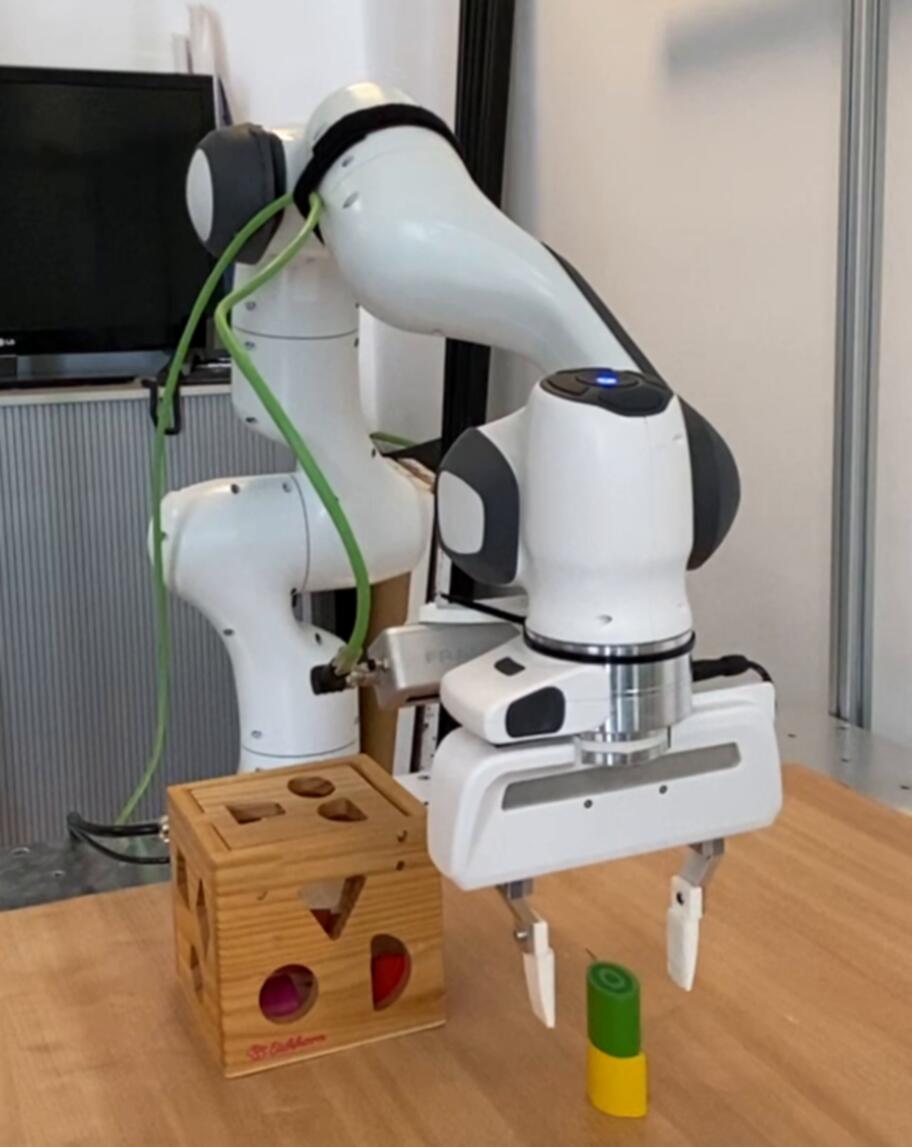}
 \includegraphics[width=.24\linewidth]{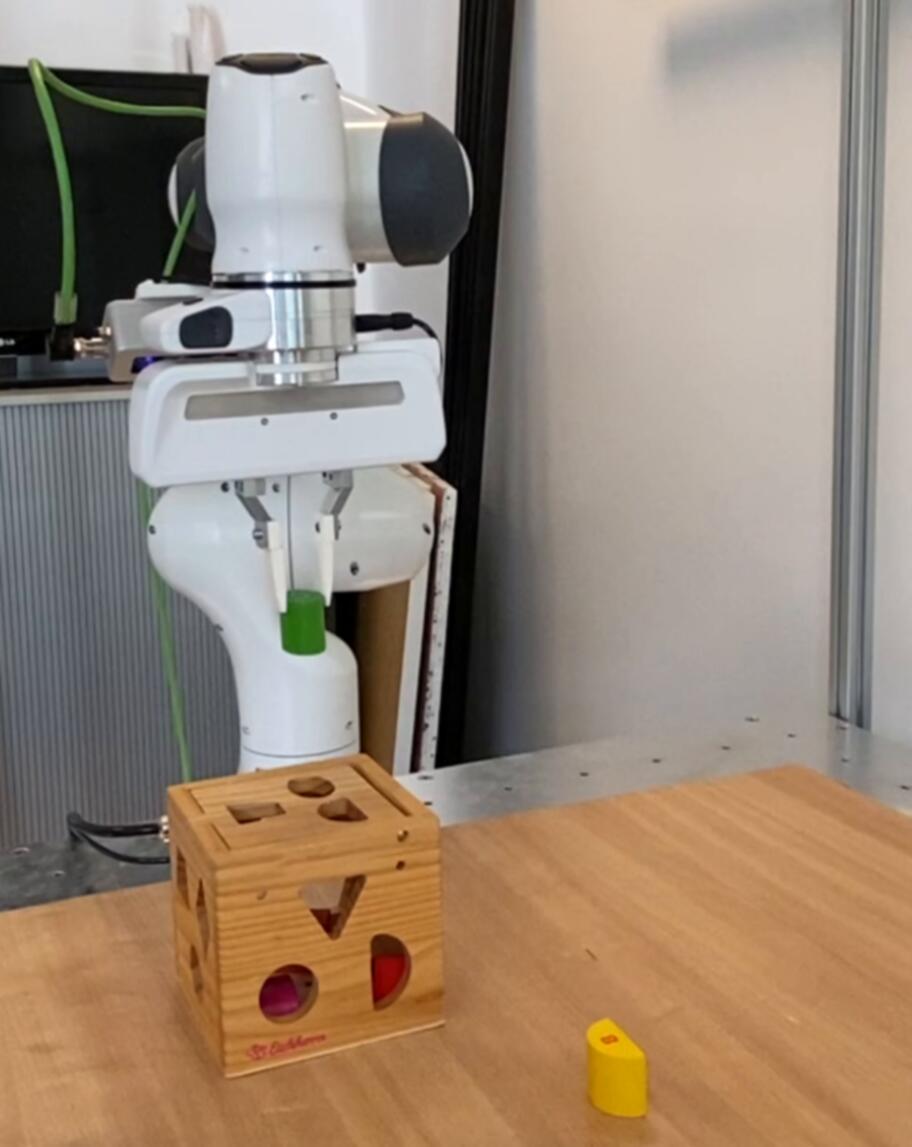}
 \includegraphics[width=.24\linewidth]{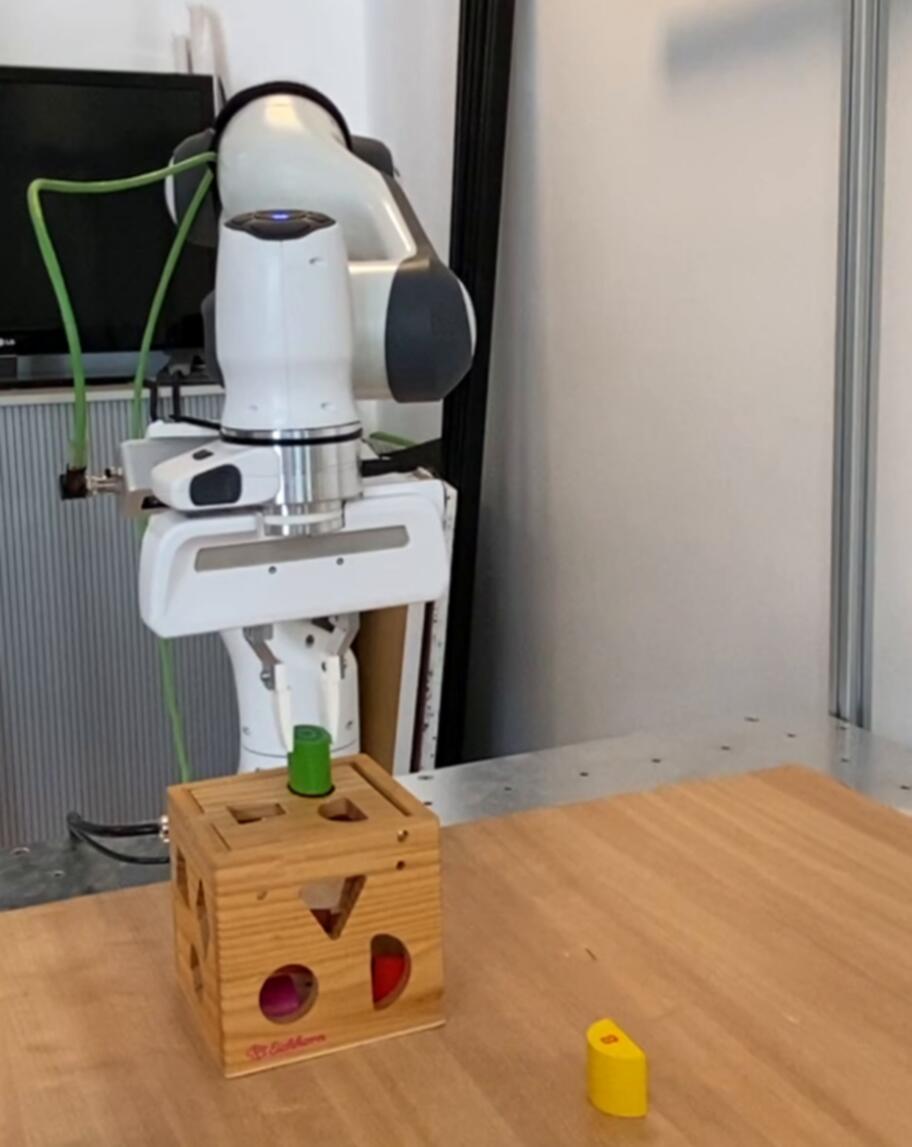}
 \includegraphics[width=.24\linewidth]{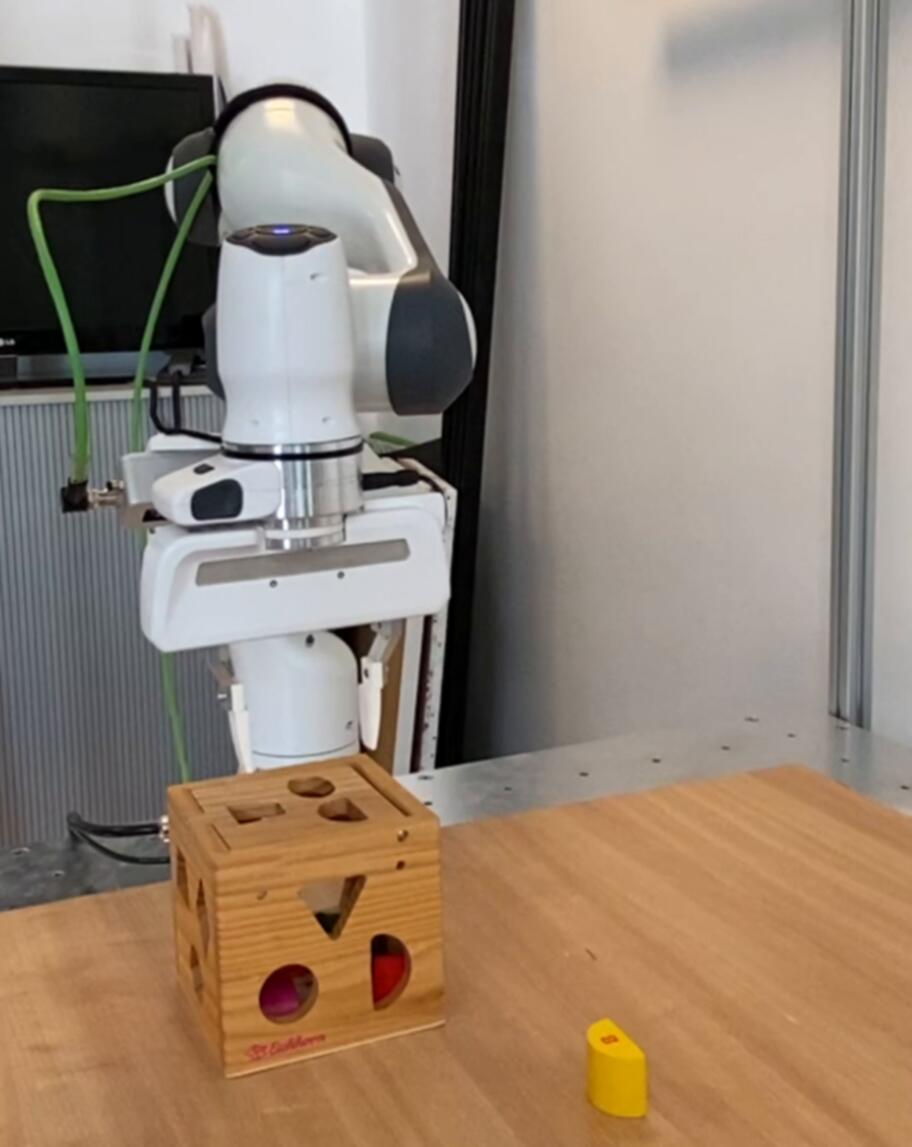}\vspace{0.5em}
 \includegraphics[width=.24\linewidth]{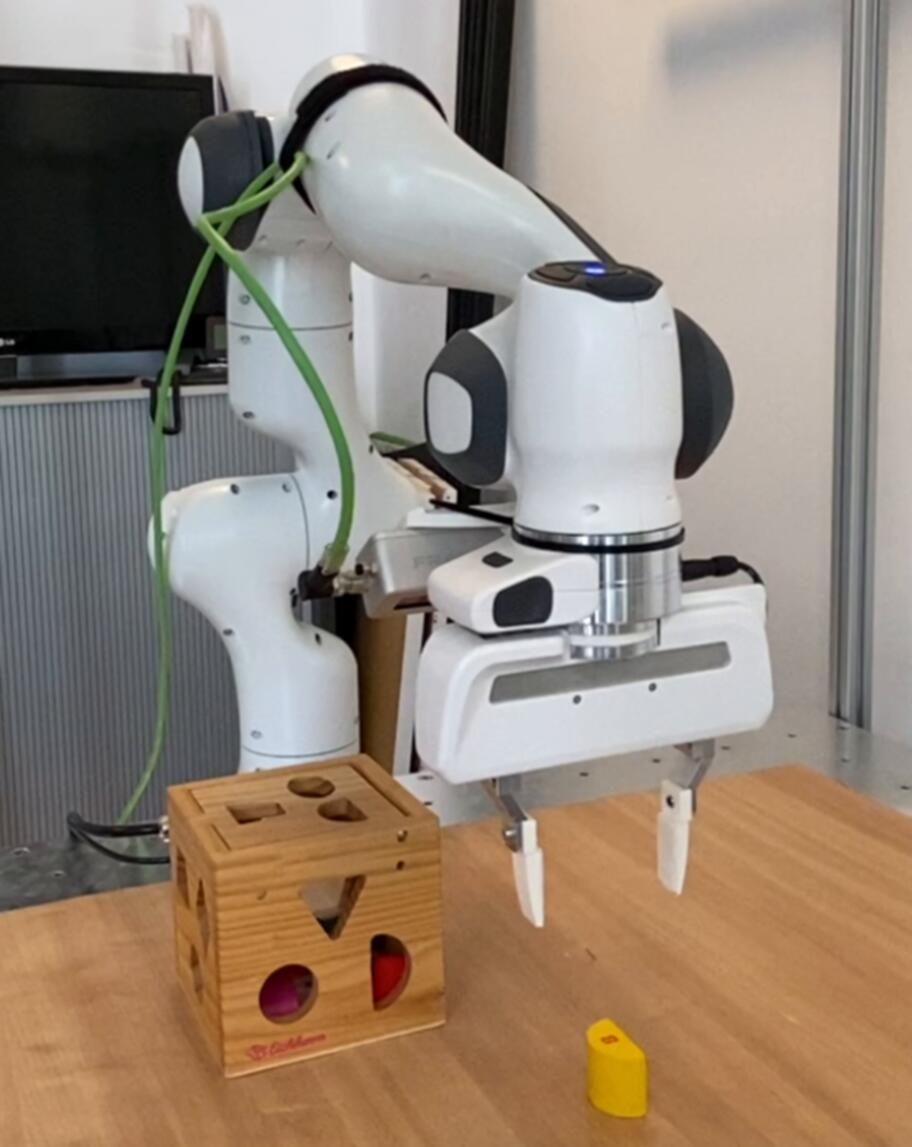}
 \includegraphics[width=.24\linewidth]{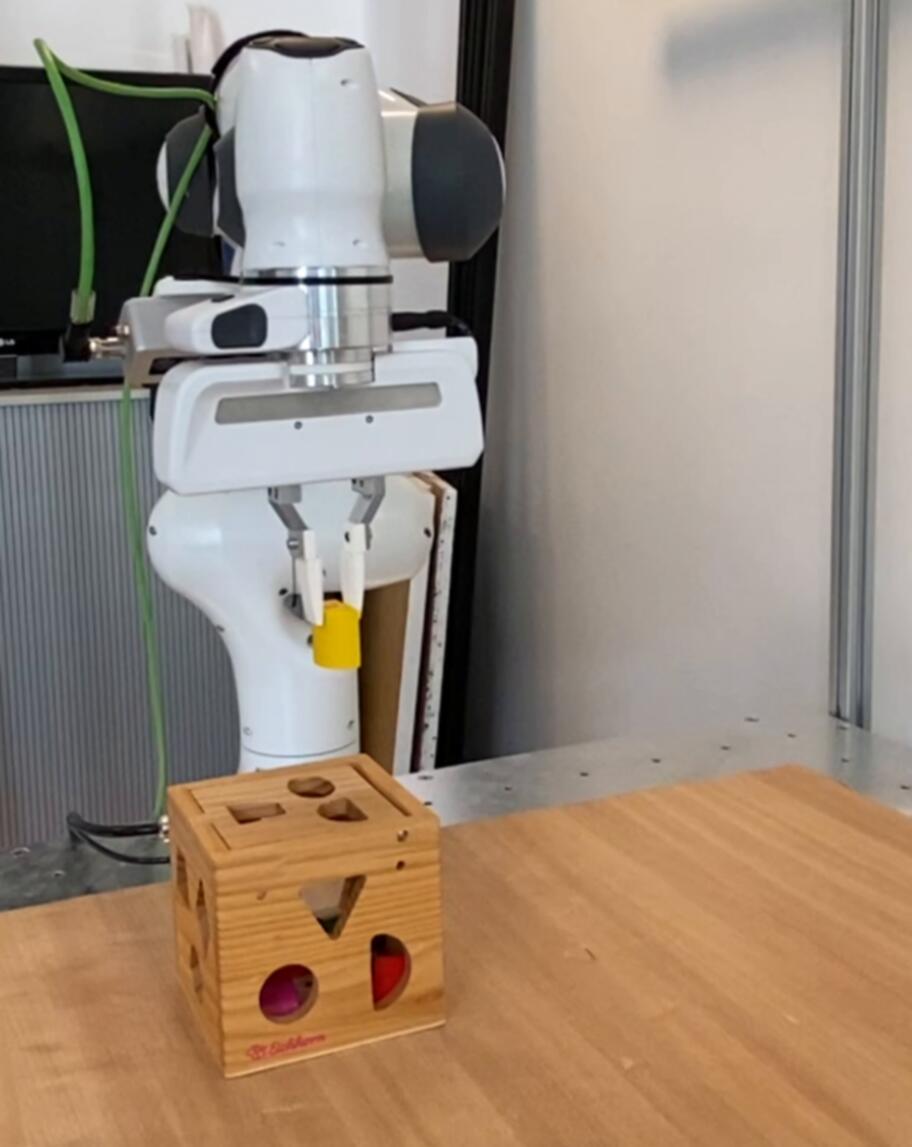}
 \includegraphics[width=.24\linewidth]{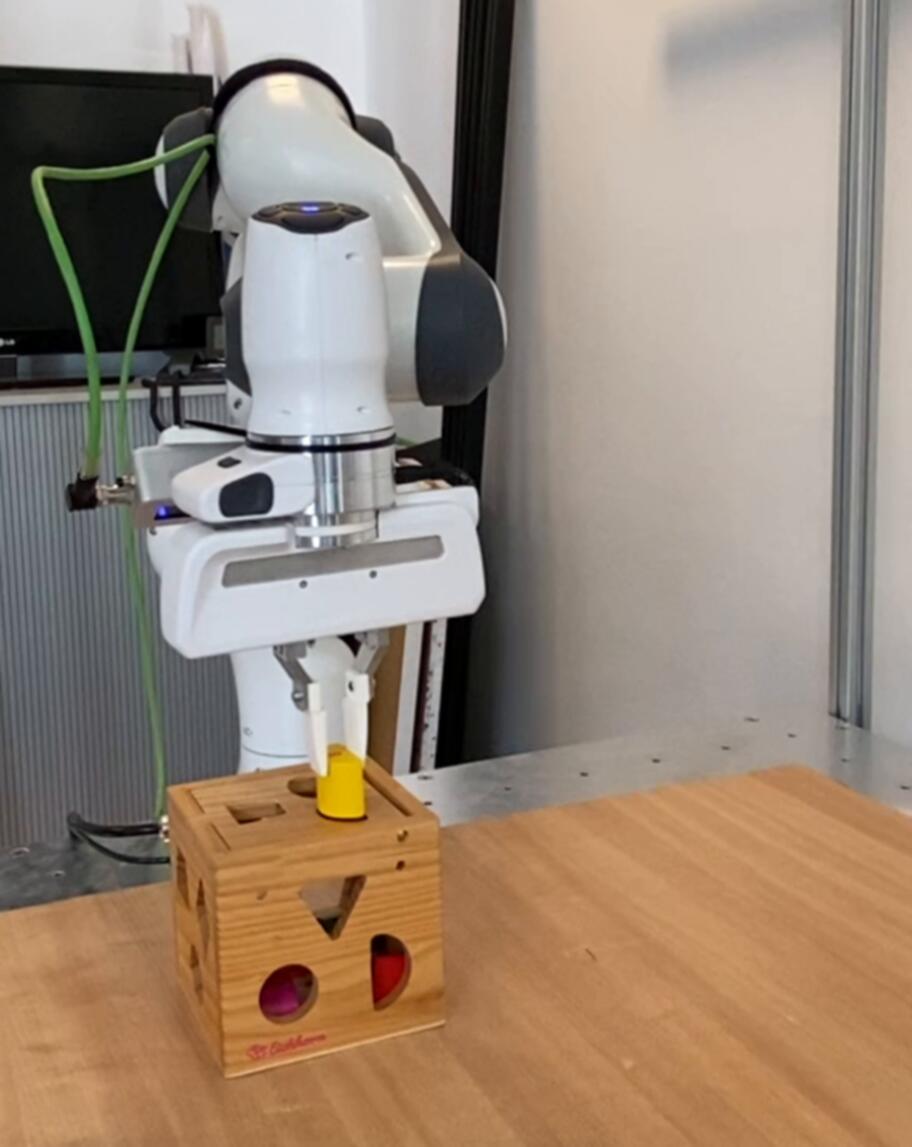}
 \includegraphics[width=.24\linewidth]{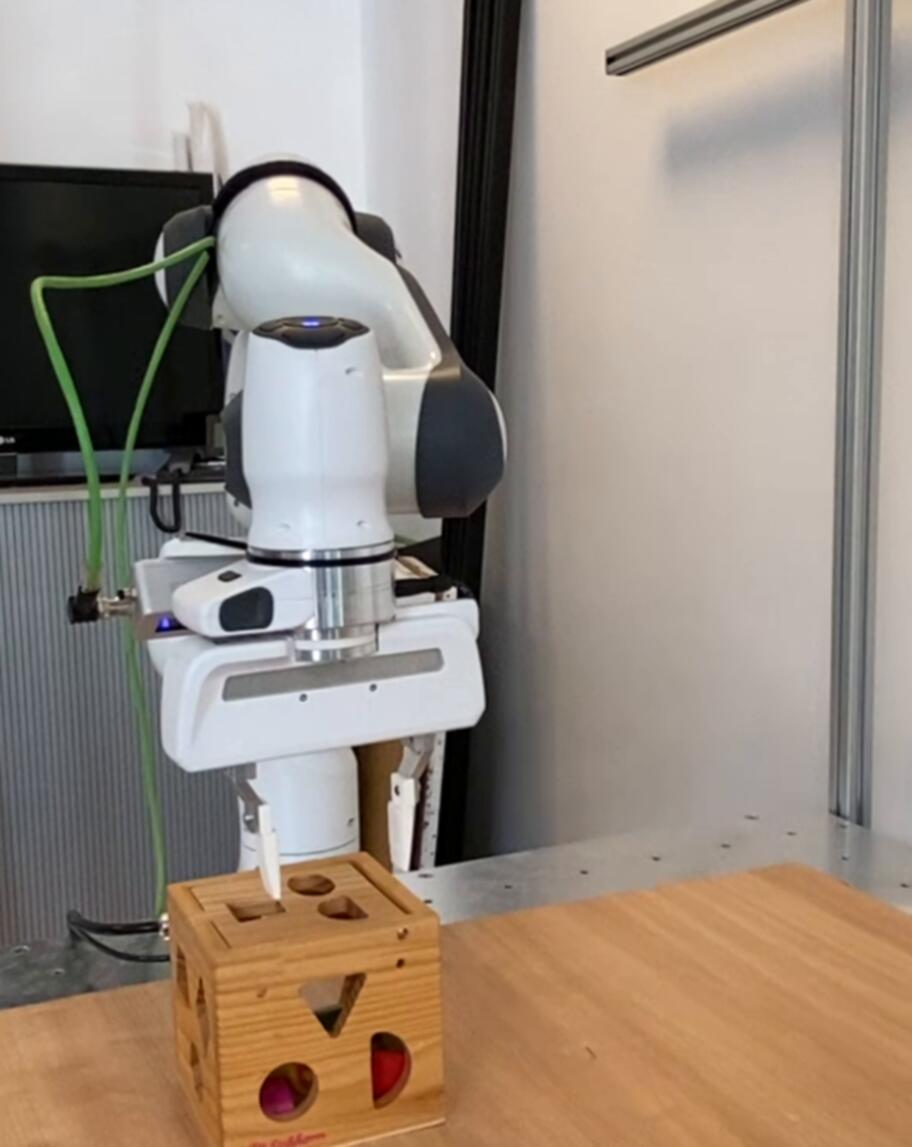}
 \vspace{.4em}
 \caption{Using a set of demonstrations allows tackling an initial configuration of stacked shapes. This shows that demonstrations allow addressing scenes that can be permuted geometrically, which would otherwise require an unfeasible number of complete demonstration sequences.}
 \label{fig:video_frames}
\end{figure}

\vspace{1em}
\textbf{Scene Permutations:}
The quantitative experiments shown to this point improve performance through improved flow and registration. The conditional approach, however, also has other advantages: it is able to deal with geometric permutations of the scene. We show this experimentally by giving the robot a challenging scenario, in which the shapes are initially stacked in a tower.
To solve this task without dividing demonstrations into subtasks one would theoretically need a demonstration for each order in which the cubes could be stacked. This is clearly infeasible for complex scenes as the number of required demonstrations to cope with all possible permutations increases rapidly. Examples from this experiment are shown in Figure \ref{fig:video_frames}.


\vspace{.5em}
\textbf{Robustness:} As an additional verification that our method does not lead to any unexpected failure cases, we evaluate a number of robustness tests. These introduce variations between the live view and the demonstration view, for example by changing the lighting conditions or the backgrounds for the tasks. Example images from these experiments are shown in Figure \ref{fig:robustness}.

\newcommand{\rbtwidth}{0.44\columnwidth}

\begin{figure}[tb]
\centering
\renewcommand{\arraystretch}{0.8}
\begin{tabular}{@{}c@{}c@{}}
    \settototalheight{\dimen1}{\includegraphics[width=\rbtwidth]{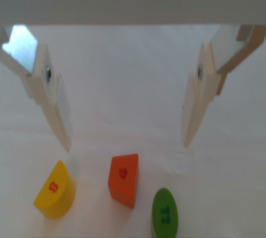}}
    \includegraphics[width=\rbtwidth]{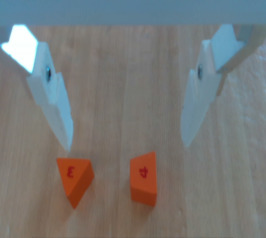}%
    \llap{\raisebox{\dimen1-\rbtwidth*\real{0.4}}{
            \includegraphics[height=\rbtwidth*\real{0.4}]{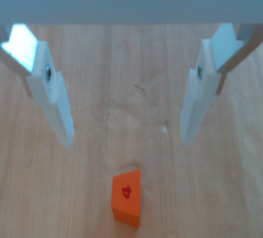}%
    }} &
    \settototalheight{\dimen2}{\includegraphics[width=\rbtwidth]{figures/robustness/background_1}}
    \includegraphics[width=\rbtwidth]{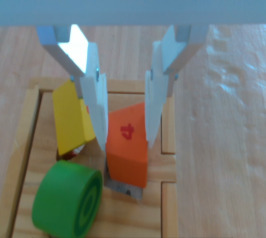}%
    \llap{\raisebox{\dimen2-\rbtwidth*\real{0.4}}{
            \includegraphics[height=\rbtwidth*\real{0.4}]{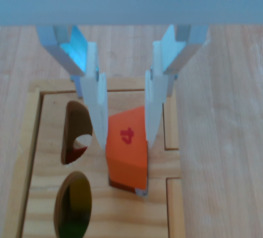}%
    }} \\
    (a) Same Color Objects & (b) Occlusions \\ & \\
    \settototalheight{\dimen3}{\includegraphics[width=\rbtwidth]{figures/robustness/background_1}}
    \includegraphics[width=\rbtwidth]{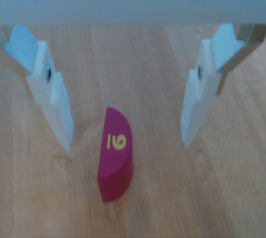}%
    \llap{\raisebox{\dimen3-\rbtwidth*\real{0.4}}{
            \includegraphics[height=\rbtwidth*\real{0.4}]{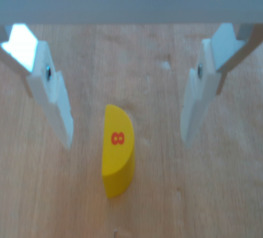}%
    }} &
    \settototalheight{\dimen4}{\includegraphics[width=\rbtwidth]{figures/robustness/background_1}}
    \includegraphics[width=\rbtwidth]{figures/robustness/background_1}%
    \llap{\raisebox{\dimen4-\rbtwidth*\real{0.4}}{
            \includegraphics[height=\rbtwidth*\real{0.4}]{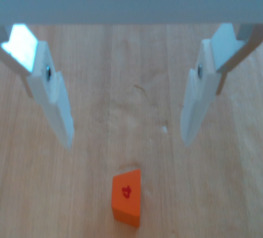}%
    }} \\
    (c) Different Color Objects & (d) New Backgrounds \\ & \\
    \settototalheight{\dimen5}{\includegraphics[width=\rbtwidth]{figures/robustness/background_1}}
    \includegraphics[width=\rbtwidth]{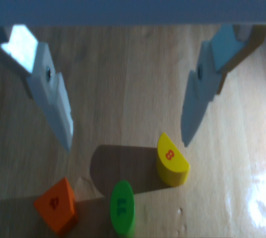}%
    \llap{\raisebox{\dimen5-\rbtwidth*\real{0.4}}{
            \includegraphics[height=\rbtwidth*\real{0.4}]{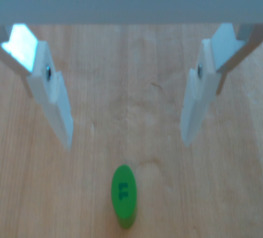}%
    }} &
    \settototalheight{\dimen6}{\includegraphics[width=\rbtwidth]{figures/robustness/background_1}}
    \includegraphics[width=\rbtwidth]{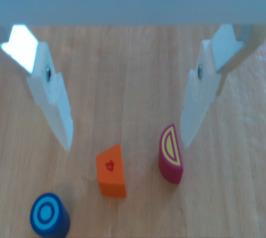}%
    \llap{\raisebox{\dimen6-\rbtwidth*\real{0.4}}{
            \includegraphics[height=\rbtwidth*\real{0.4}]{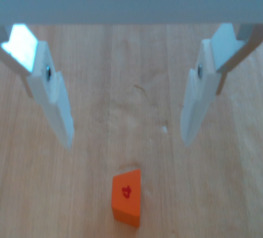}%
    }} \\
    (e) Lighting & (f) Unseen Objects \\
\end{tabular}

        \caption{Robustness to a number of nuisance variations results from the use of learned optical flow. Large images show live views and inset images show demonstrations. All examples from recording of successful insertions.}
        \label{fig:robustness}
\renewcommand{\arraystretch}{1.0}
\end{figure}

\section{CONCLUSIONS}


We have presented conditional servoing as a concept -- an outer loop around a Servoing-from-Demonstration loop. A practical implementation based on the FlowControl method has been provided. This implementation was tested both in simulation and on a real robot. In these experiments, we showed that reprojection error is a suitable scoring function for demonstration sequence selection. The method achieves high success rates on the challenging problem solving an insertion cube. In addition to these quantitative tests, we conducted qualitative tests showing that the system is able to handle scene permutations and works robustly despite the presence of nuisance variations in the scene.

\section*{Acknowledgments}
This work was partially funded by the German Research Foundation (DFG) under project number BR 3815/10-1.

\vspace{1em}

\FloatBarrier
\clearpage
\newpage

\bibliographystyle{IEEEtran}
\bibliography{references}

\end{document}